\documentclass{article}
\usepackage{spconf,amsmath,graphicx,hyperref}

\usepackage{amsfonts}
\usepackage{booktabs}
\usepackage{multirow}

\usepackage{cite}
\usepackage{amsmath}
\usepackage{amsfonts}
\usepackage{graphicx} 
\usepackage{csvsimple,booktabs}
\usepackage{float,enumitem}
\usepackage{graphicx,epsfig,xcolor}
\usepackage{balance}
\usepackage{caption}
\usepackage[caption=false,font=footnotesize]{subfig}

\def\BibTeX{{\rm B\kern-.05em{\sc i\kern-.025em b}\kern-.08em
    T\kern-.1667em\lower.7ex\hbox{E}\kern-.125emX}}

\title{MS-DFTVNet:A Long-Term Time Series Prediction Method Based on Multi-Scale Deformable Convolution}
\name{
Chenghan Li$^{1,*}$ \quad
Mingchen Li$^{2,*}$ \quad
Yipu Liao$^{3}$ \quad
Ruisheng Diao$^{4,\dagger}$
}

\address{
$^{1}$ Shenzhen International Graduate School, Tsinghua University, Shenzhen, China \\
$^{2}$ The Hong Kong University of Science and Technology (Guangzhou), Guangzhou, China \\
$^{3}$ Department of EECS, University of Michigan, Ann Arbor, USA \\
$^{4}$ ZJU-UIUC Institute, Zhejiang University, Haining, China \\
Email: \texttt{ruishengdiao@intl.zju.edu.cn} \\
$^{*}$ Equal contribution \quad $^{\dagger}$ Corresponding author
}

\begin{document}
%
\maketitle
\begin{abstract}
Research on long-term time series prediction has primarily relied on Transformer and MLP models, while the potential of convolutional networks in this domain remains underexplored. To address this, we propose a novel multi-scale time series reshape module that effectively captures cross-period patch interactions and variable dependencies. Building on this, we develop MS-DFTVNet, the multi-scale 3D deformable convolutional framework tailored for long-term forecasting. Moreover, to handle the inherently uneven distribution of temporal features, we introduce a context-aware dynamic deformable convolution mechanism, which further enhances the model’s ability to capture complex temporal patterns. Extensive experiments demonstrate that MS-DFTVNet not only significantly outperforms strong baselines but also achieves an average improvement of about 7.5\% across six public datasets, setting new state-of-the-art results.
\end{abstract}

\section{Introduction}
\begin{figure}[!t]
    \centering
    \subfloat[Comparison of standard, dilated, and deformable convolutions]{%
        \includegraphics[width=0.48\columnwidth,keepaspectratio]{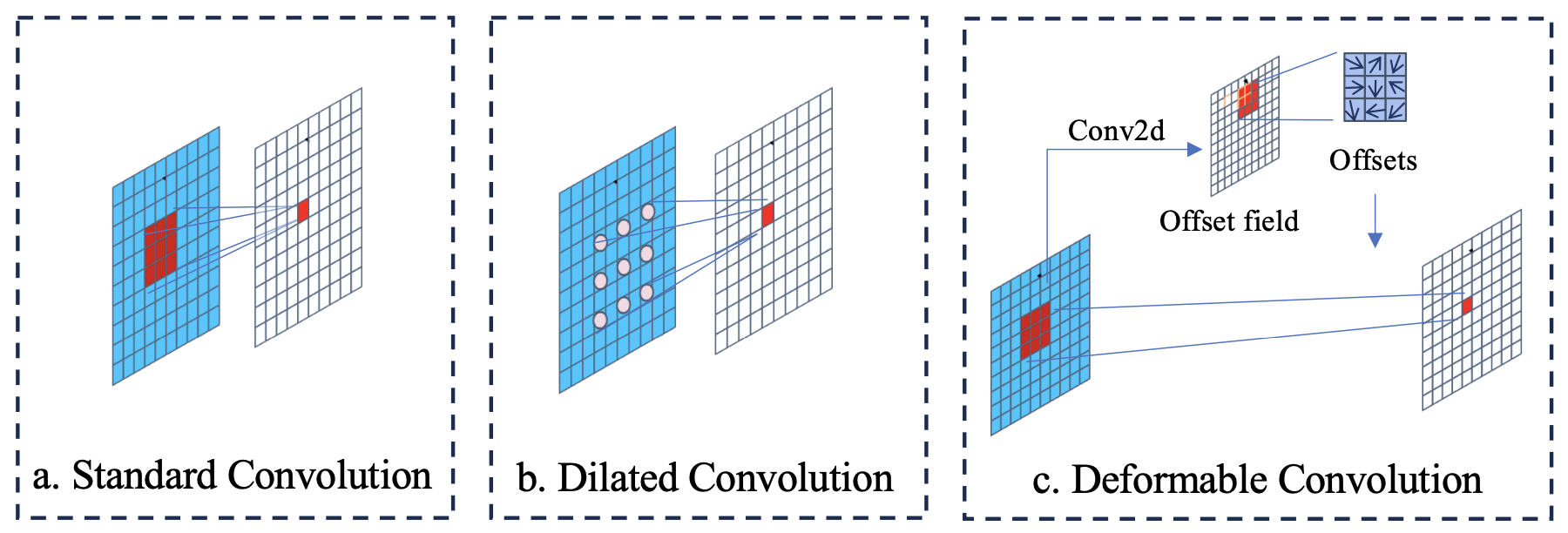}%
        \label{fig:sub1}}
    \hfil
    \subfloat[Multi-scale time series with correlations]{%
        \includegraphics[width=0.48\columnwidth,keepaspectratio,page=1]{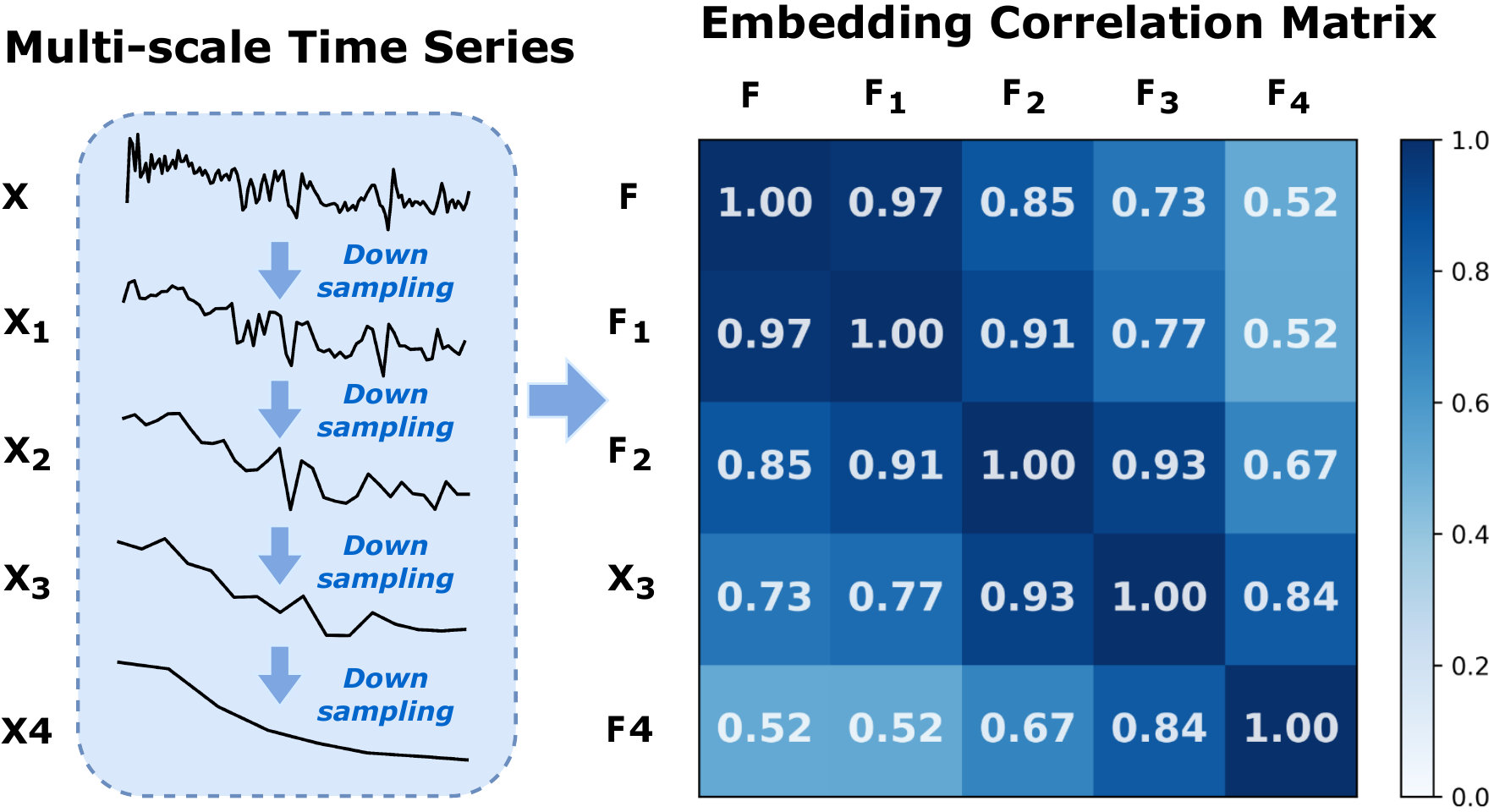}%
        \label{fig:sub2}}
    \caption{Illustration of deformable convolution and multi-scale characteristics in time series data}
    \label{fig:intro}
\end{figure}

Long-term time series prediction is a crucial task with extremely wide applications in professional fields such as energy management\cite{nivolianiti2024energy}\cite{dong2024artificial}, weather forecasting\cite{price2025probabilistic}, and health monitoring\cite{yu2024implantable}. However, long-term time series forecasting tasks often involve complex multi-period scales\cite{huang2024fl}\cite{song2024multi} and cross-variable dependencies\cite{litvnet}, making the development of effective forecasting models a significant challenge. Time series data can be partitioned into segments of different lengths, where smaller segments provide fine-grained temporal features and larger segments capture coarse-grained temporal patterns. Local dependencies are typically reflected in short-range temporal relationships, while global dependencies arise from long-range temporal interactions \cite{chen2023long}\cite{yi2023frequency}\cite{cai2024msgnet}(see in Figure \ref{fig:intro}). Recent advances in convolutional models have demonstrated promising progress in long-term time-series forecasting. TimesNet\cite{wu2022timesnet}, ModernTCN\cite{luo2024moderntcn} extend convolutional approaches to capture complex temporal dynamics and global features. However, these models still fail to fully exploit the multi-scale characteristics of time series and the direct relationships across variables. Multi-scale feature modeling has proven highly effective for correlation learning and feature extraction in domains such as computer vision and multimodal learning. Yet, its potential remains underexplored in time series forecasting. Methods like Pyraformer\cite{huang2025improving}, Scaleformer\cite{shabani2022scaleformer} introduce multi-scale frameworks, but often at the cost of added model complexity. Despite these advances, existing methods still face fundamental challenges in effectively modeling the intrinsic properties of time series. In particular, they struggle to adaptively capture multi-scale temporal features while simultaneously modeling direct dependencies across variables. These limitations motivate our work.

\begin{figure*}[ht]
\centering
\includegraphics[width=0.85\textwidth]{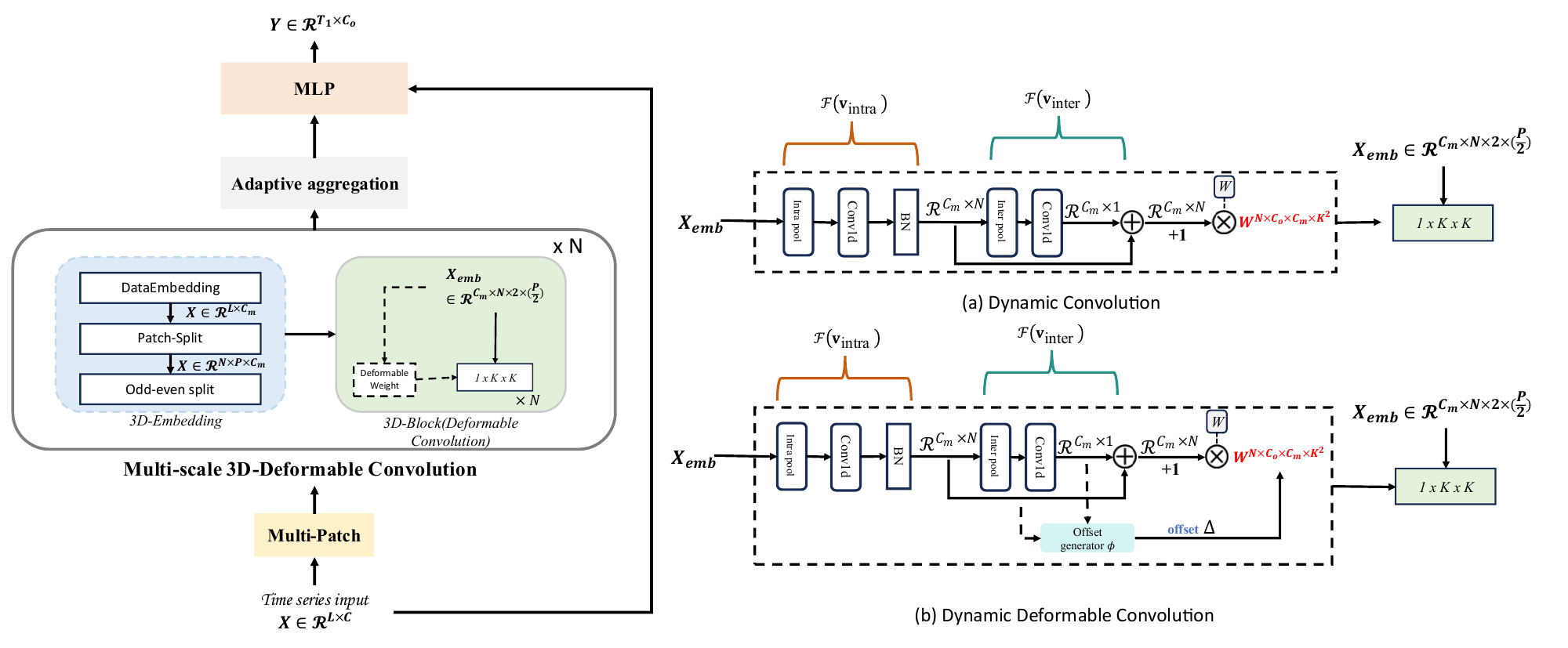}
\caption{Overall framework of MS-DFTVNet with the dynamic deformable convolution module.}
\label{fig:method}
\end{figure*}
We address two major limitations in long-term time-series forecasting: (1) the lack of adaptive multi-scale feature extraction in existing models, and (2) the absence of CNN-based methods that can jointly capture multi-scale patches and cross-variable dependencies. Motivated by our prior work~\cite{litvnet}, which showed that reshaping time-series patches into 3D tensors with dynamic convolution can effectively model both time-varying and time-invariant patterns, we introduce MS-DFTVNet, an adaptive multi-scale 3D dynamic deformable convolutional network. Our key contributions are: (1) a Fourier-based adaptive patch module that enables efficient multi-scale feature extraction, and (2) a dynamic deformable convolution mechanism that captures both patch-level structures and cross-variable dependencies. Extensive evaluations on multiple public benchmarks demonstrate that MS-DFTVNet delivers consistent gains in accuracy and robustness, establishing new state-of-the-art performance.


\section{Proposed Method}
\subsection{Problem Statement}
Given a multivariate history \(X_{\mathrm{in}} \in \mathbb{R}^{L \times C}\) of length \(L\), the goal is to forecast the future sequence \(X_{\mathrm{out}} \in \mathbb{R}^{P \times C}\) with horizon \(P\). Let the predictor be \(f_\theta\), i.e., \(\hat{X}_{\mathrm{out}} = f_\theta(X_{\mathrm{in}})\). The parameters \(\theta\) are learned by minimizing the supervised loss \(\mathcal{L}(\theta) = \tfrac{1}{PC}\|X_{\mathrm{out}}-\hat{X}_{\mathrm{out}}\|_2^2\), where mean squared error (MSE) is used by default.

\subsection{Framework}
The proposed method is illustrated in Figure~\ref{fig:method}. The overall architecture of MS-DFTVNet is based on a Mixer-based framework and is composed of three key modules: \textit{Multi-Patch}, \textit{3D Deformable Convolution}, and \textit{Adaptive Aggregation}. Each of these modules plays a distinct role in enhancing Multi-Scale feature extraction and fusion. The \textit{Multi-Patch} module is designed to capture local and global temporal dependencies across multiple time scales and the \textit{3D Deformable Convolution} module dynamically adjusts its receptive field to model complex multi-scale temporal interactions more effectively. Finally, the \textit{Adaptive Aggregation} module integrates features from multiple paths to produce time series representations. In the following sections, we provide a detailed discussion of each component and its contribution to the model's overall performance.

\textit{Multi-Patch Module}:
This module utilizes the Fast Fourier transform to extract multiple dominant periods embedded within the time series adaptively. By identifying these intrinsic periodic patterns, the time series can be segmented into patches that align with the discovered periods, enabling better structural representation and improved modeling efficiency. For a time series of length \(L\) with \(C\) features, the original format is represented as \(\mathcal{X}_{\text{in}} \in \mathbb{R}^{L \times C},\)
where each row corresponds to a specific timestamp and each column denotes a particular variable. To effectively capture temporal dependencies across varying scales, it is crucial first to identify the underlying periodic components. This is achieved by transforming the time series into the frequency domain using the Fast Fourier Transform (FFT), which enables the detection of frequency peaks corresponding to dominant periodicities. The resulting spectral information serves as the foundation for adaptive patch segmentation, as detailed below:
\begin{equation}
\begin{aligned}
\{f_1, \cdots, f_k\} &= \underset{f_* \in \{1, \cdots, \left\lfloor \frac{T}{2} \right\rfloor\}}{\arg \text{Topk}} (\mathbf{\text{Avg}\left(\text{Amp}\left(\text{FFT}(\mathcal{X}_{\text{in}})\right)\right)})\\
p_i &= \left\lfloor \frac{T}{f_i} \right\rfloor, \quad i \in \{1, \cdots, k\}
\end{aligned}
\end{equation}
where \( \text{FFT}(\cdot) \) and \( \text{Amp}(\cdot) \) represent the Fast Fourier Transform and the calculation of amplitude values, respectively. The amplitude of each frequency is computed by averaging across \( C \) dimensions using \( \text{Avg}(\cdot) \). This corresponds to the patch length \( p_i = \frac{T}{f_i} \), where \( f_i \) denotes the frequency.

\textit{3D Deformable Convolution}:
We first embed the input time series along the feature dimension, mapping it from \(C\) to a high-dimensional space of \(C_m\), resulting in the embedded representation \(\mathcal{X}_{\text{in}} \in \mathbb{R}^{L \times C_m}\). This step enhances the feature expressiveness. To segment the sequence into temporal patches($p_i$), we apply a one-dimensional convolutional layer (Conv1D) with a kernel size of \(P\), effectively dividing \(\mathcal{X}_{\text{in}}\) into \(N = L/P\) patches. The output after convolution is denoted as \(\mathcal{X}_{\text{emb}} \in \mathbb{R}^{N \times P \times C_m}\), where each patch preserves local temporal patterns over \(P\) time steps. Next, to further capture intra-patch structure, we split each patch into two equal-length sub-patches of size \(P/2\), and then stack them along a new dimension, yielding the reshaped representation \(\mathcal{X}_{\text{emb}} \in \mathbb{R}^{N \times 2 \times (P/2) \times C_m}\). This formulation enables the model to learn more nuanced dependencies within each patch. When using multiple patch lengths \(p_i\), we can obtain a set of multi-scale embeddings \(\{\mathcal{X}^1_{\text{emb}}, \ldots, \mathcal{X}^k_{\text{emb}}\}\), enabling the model to capture dynamics at different temporal resolutions.

Building upon TVNet, which models dynamic amplitude modulation through $W_i = \alpha_i \cdot W_b$, we extend this formulation by additionally learning kernel \textit{sampling shapes}. Specifically, for each scale $p_i$ and patch $n$, continuous offsets $\{\Delta^{(i)}_{n,k}\}$ are generated from intra-patch and inter-patch vectors, and bounded to ensure scale-aware ranges. The output feature is then expressed as:
\begin{equation}
\tilde{x}^{(i)}_{n} = \sum_{k \in \mathcal{K}} \big(\textcolor{red}{\alpha^{(i)}_{n,k}}\, W_{b,k}\big)\cdot \textcolor{blue}{\Delta^{(i)}_{n,k}}
\end{equation}
where $\mathcal{K}$ denotes kernel locations.

The weight $\alpha^{(i)}$ is generated adaptively by considering both intra-patch and inter-patch vectors, ensuring that local and global dynamics are jointly captured:
\begin{equation}
\alpha^{(i)} = 1 + \mathcal{F}_{\text{intra}}(\mathbf{v}_{\text{intra}}) + \mathcal{F}_{\text{inter}}(\mathbf{v}_{\text{inter}})
\end{equation}
where $\mathbf{v}_{\text{intra}}$ and $\mathbf{v}_{\text{inter}}$ are obtained by hierarchical pooling, followed by a lightweight Conv1D layer and ReLU activations. Beyond amplitude modulation, we further introduce shape dynamics by learning kernel sampling offsets. Specifically, offsets $\{\Delta^{(i)}_{n,k}\}$ are generated by an offset generator that reuses the same intra- and inter-patch vectors, i.e.,
\begin{equation}
\widehat{\Delta}^{(i)}_n\!\big(\{\widehat{\Delta}^{(i)}_{n,k}\}_{k \in \mathcal{K}}\big)
= \Psi\!\big(\mathbf{v}^{(i)}_{\text{intra}},\, \mathbf{v}^{(i)}_{\text{inter}}\big)
\end{equation}
\noindent where $\Psi(\cdot,\cdot)$ denotes the offset generator. A lightweight head (e.g., $1\times1$ convolution with grouped linear layers) maps vectors to offsets. To stabilize training, offsets are bounded using a scaled $\tanh$:
\begin{equation}
\Delta^{(i)}_{n,k} = r_t(p_i)\,\tanh\!\big(\widehat{\Delta}{}^{(i)}_{n,k}\big)
\end{equation}
Empirically, we set $r_t(p_i) = \lfloor p_i/4 \rfloor$, enabling larger receptive shifts for longer periods.


\textit{Adaptive aggregation}:involves fusing \( k \) distinct 3D-feature vectors \( \{ \mathcal{\hat{X}}^1_{\text{emb}}, \ldots, \mathcal{\hat{X}}^k_{\text{emb}} \} \) for the subsequent layer. The amplitudes \( \mathcal{A} \) reflect the relative importance of selected frequencies and periods, corresponding to the importance of each transformed 3D tensor. We aggregate the 1D-representations based on these amplitudes:
\begin{subequations}
\begin{align}
\mathcal{A}_{f_1}, \cdots, \mathcal{A}_{f_k} &= \arg \text{Topk} \left( \text{Avg} \left( \text{Amp} \left( \text{FFT}(\mathcal{X}_{\text{in}}) \right) \right) \right) \label{eq:topk} \\
\hat{\mathcal{A}}_{f_1}, \cdots, \hat{\mathcal{A}}_{f_k} &= \text{Softmax} \left( \mathcal{A}_{f_1}, \cdots, \mathcal{A}_{f_k} \right) \label{eq:softmax} \\
\mathcal{X}_{\text{emb}} &= \sum_{i=1}^{k} \hat{\mathcal{A}}_{f_i} \times \mathcal{\hat{X}}^{i}_{\text{emb}}. \label{eq:aggregation}
\end{align}
\end{subequations}

The final prediction results are obtained using residual connections and an MLP, as shown in the following formulas:
\begin{equation}
Y=MLP(\mathcal{X}_{emb}+\mathcal{X}_{in}).
\end{equation}

\section{Experiment and Result}
In this section, we evaluate the performance of MS-DFTVNet on public datasets using Mean Squared Error (MSE) and Mean Absolute Error (MAE) as our metrics. It is worth noting that, for these metrics, a lower value indicates a superior performance. To fully validate the effectiveness of the proposed method in this paper, we conducted comparisons on six public datasets (Exchange, ETTm2, ETTh1, Electricity, Weather, ILI) with nine baselines( TVNet\cite{litvnet}, PatchTST\cite{yi2024patchesnet}, iTransformer\cite{liu2023itransformer}, Crossformer\cite{zhang2023crossformer}, RLinear\cite{li2023revisiting}, MTS-Mixer\cite{li2023mts}, DLinear, TimesNet\cite{wu2022timesnet}, and MICN\cite{wang2023micn}). Our method and baselines are optimized using the Adam optimizer. The learning rate starts at \(1e^{-4}\) and is halved when loss stagnates. The loss function used is MSE. We evaluate accuracy using \(MSE = \frac{1}{n} \sum_{i=1}^{n} (y - \hat{y})^2\) and \(MAE = \frac{1}{n} \sum_{i=1}^{n} |y - \hat{y}|\). All models are run on a single NVIDIA 3090 GPU with 24 GB of memory.

\begin{table*}[!t]
\setlength{\tabcolsep}{1pt}
 \centering
 \caption{Forecasting result comparison with different horizon lengths. The lookback length is set to 24 for ILI and 96 for the others. \textcolor{red}{Red} indicates the best result, while \textcolor{blue}{Blue underlining} indicates the second-best result.}
 \label{tab:res}
 \newcommand{\tabincell}[2]{\begin{tabular}{@{}#1@{}}#2\end{tabular}}
 \scalebox{0.85}{
 \begin{tabular}{c|c|cccccccccccccccccccc}
  \toprule
  \multicolumn{2}{c}{Models} & \multicolumn{2}{c}{\textcolor{red}{MS-DFTVNet}} & \multicolumn{2}{c}{TVNet} & \multicolumn{2}{c}{PatchTST} & \multicolumn{2}{c}{iTransformer} & \multicolumn{2}{c}{Crossformer} & \multicolumn{2}{c}{RLinear} & \multicolumn{2}{c}{MTS-Mixer} & \multicolumn{2}{c}{DLinear} & \multicolumn{2}{c}{TimesNet} & \multicolumn{2}{c}{MICN} \\
  \cmidrule(r){3-4} \cmidrule(r){5-6} \cmidrule(r){7-8} \cmidrule(r){9-10} \cmidrule(r){11-12} \cmidrule(r){13-14} \cmidrule(r){15-16} \cmidrule(r){17-18} \cmidrule(r){19-20} \cmidrule(r){21-22}
  \multicolumn{2}{c}{Metrics} & MSE & MAE & MSE & MAE & MSE & MAE & MSE & MAE & MSE & MAE & MSE & MAE & MSE & MAE & MSE & MAE & MSE & MAE & MSE & MAE  \\ 
  \midrule 
  \multirow{4}{*}{\rotatebox{90}{Exchange}} & 96 & \textcolor{blue}{\underline{0.081}} & \textcolor{blue}{\underline{0.203}} & \textcolor{red}{0.080} & \textcolor{red}{0.195} & 0.093 & 0.214 & {0.086} & {0.206} & 0.186 & 0.346 & 0.083 & 0.301 & 0.083 & 0.200 & \textcolor{blue}{\underline{0.081}} & \textcolor{blue}{\underline{0.203}} & 0.107 & 0.234 & 0.102 & 0.235 \\
  & 192 & \textcolor{red}{0.157} & \textcolor{blue}{\underline{0.289}} & \textcolor{blue}{\underline{0.163}} & \textcolor{red}{0.285} & 0.192 & 0.312 & 0.177 & 0.299 & 0.467 & 0.522 & 0.170 & 0.293 & 0.174 & 0.296 & \textcolor{red}{0.157} & 0.293 & 0.226 & 0.344 & 0.172 & 0.316 \\
  & 336 & \textcolor{red}{0.271} & \textcolor{red}{0.390} & \textcolor{blue}{\underline{0.291}} & \textcolor{blue}{\underline{0.394}} & 0.350 & 0.432 & 0.331 & 0.417 & 0.783 & 0.721 & 0.309 & 0.401 & 0.336 & 0.417 & 0.305 & 0.414 & 0.367 & 0.448 & \textcolor{blue}{\underline{0.272}} & 0.407 \\
  & 720 & \textcolor{red}{0.642} & \textcolor{blue}{\underline{0.601}} & 0.658 & \textcolor{red}{0.594} & 0.911 & 0.716 & 0.847 & 0.691 & 1.367 & 0.943 & 0.817 & 0.680 & 0.900 & 0.715 & \textcolor{blue}{\underline{0.643}} & 0.601 & 0.964 & 0.746 & 0.714 & 0.658 \\
  \midrule 
  \multirow{4}{*}{\rotatebox{90}{ETTm2}} & 96 & \textcolor{red}{0.134} & \textcolor{red}{0.249} & \textcolor{blue}{\underline{0.161}} & \textcolor{blue}{\underline{0.254}} & 0.165 & 0.255 & 0.180 & 0.264 & 0.421 & 0.461 & 0.164 & 0.253 & 0.177 & 0.259 & 0.167 & 0.260 & 0.187 & 0.267 & 0.178 & 0.273 \\
  & 192 & \textcolor{red}{0.201} & \textcolor{red}{0.282} & 0.220 & 0.293 & 0.220 & 0.292 & 0.250 & 0.309 & 0.503 & 0.519 & 0.219 & 0.290 & 0.241 & 0.303 & 0.224 & 0.303 & 0.249 & 0.309 & 0.245 & 0.316 \\
  & 336 & \textcolor{red}{0.263} & \textcolor{red}{0.313} & \textcolor{blue}{\underline{0.272}} & \textcolor{blue}{\underline{0.316}} & 0.274 & 0.329 & 0.311 & 0.348 & 0.611 & 0.580 & 0.273 & 0.326 & 0.297 & 0.338 & 0.281 & 0.342 & 0.312 & 0.351 & 0.295 & 0.350 \\
  & 720 & \textcolor{red}{0.343} & \textcolor{red}{0.366} & \textcolor{blue}{\underline{0.349}} & \textcolor{blue}{\underline{0.379}} & 0.362 & 0.385 & 0.412 & 0.407 & 0.996 & 0.750 & 0.366 & 0.385 & 0.396 & 0.398 & 0.397 & 0.421 & 0.497 & 0.403 & 0.389 & 0.406 \\
  \midrule 
  \multirow{4}{*}{\rotatebox{90}{ETTh1}} & 96 & \textcolor{red}{0.354} & \textcolor{red}{0.392} & 0.371 & 0.408 & \textcolor{blue}{\underline{0.370}} & \textcolor{blue}{\underline{0.399}} & 0.386 & 0.405 & 0.386 & 0.429 & 0.366 & 0.391 & 0.372 & 0.395 & 0.375 & 0.399 & 0.384 & 0.402 & 0.396 & 0.427 \\
  & 192 & \textcolor{red}{0.390} & \textcolor{red}{0.404} & \textcolor{blue}{\underline{0.398}} & \textcolor{blue}{\underline{0.409}} & 0.413 & 0.421 & 0.441 & 0.436 & 0.419 & 0.444 & 0.404 & 0.412 & 0.416 & 0.426 & 0.405 & 0.416 & 0.557 & 0.436 & 0.430 & 0.453 \\
  & 336 & \textcolor{red}{0.399} & \textcolor{red}{0.407} & \textcolor{blue}{\underline{0.401}} & \textcolor{blue}{\underline{0.409}} & 0.422 & 0.436 & 0.487 & 0.458 & 0.440 & 0.461 & 0.420 & 0.423 & 0.455 & 0.449 & 0.439 & 0.443 & 0.491 & 0.469 & 0.433 & 0.458 \\
  & 720 & \textcolor{red}{0.437} & \textcolor{red}{0.446} & 0.458 & \textcolor{blue}{\underline{0.459}} & \textcolor{blue}{\underline{0.447}} & 0.466 & 0.503 & 0.491 & 0.519 & 0.524 & 0.442 & 0.456 & 0.475 & 0.472 & 0.472 & 0.490 & 0.521 & 0.500 & 0.474 & 0.508 \\
  \midrule 
  \multirow{4}{*}{\rotatebox{90}{Electricity}} & 96 & \textcolor{blue}{\underline{0.134}} & \textcolor{red}{0.221}  & 0.142 & 0.223 & \textcolor{red}{0.129} & \textcolor{blue}{\underline{0.222}} & 0.148 & 0.240 & 0.187 & 0.283 & 0.140 & 0.235 & 0.141 & 0.243 & 0.153 & 0.237 & 0.168 & 0.272 & 0.159 & 0.267 \\
  & 192 & 0.154 & \textcolor{red}{0.236} & 0.165 & 0.241 & \textcolor{red}{0.147} & \textcolor{blue}{\underline{0.240}} & 0.162 & 0.253 & 0.258 & 0.330 & 0.154 & 0.248 & 0.163 & 0.261 & 0.152 & 0.249 & 0.184 & 0.289 & 0.168 & 0.279 \\
  & 336 & \textcolor{red}{0.159} & \textcolor{blue}{\underline{0.260}} & 0.164 & 0.269 & \textcolor{blue}{\underline{0.163}} & \textcolor{red}{0.259} & 0.178 & 0.269 & 0.323 & 0.369 & 0.171 & 0.264 & 0.176 & 0.277 & 0.169 & 0.267 & 0.198 & 0.308 & 0.161 & 0.259 \\
  & 720 & \textcolor{red}{0.181} & \textcolor{red}{0.283} & \textcolor{blue}{\underline{0.190}} & \textcolor{blue}{\underline{0.284}} & 0.197 & 0.290 & 0.225 & 0.317 & 0.404 & 0.423 & 0.209 & 0.297 & 0.212 & 0.308 & 0.233 & 0.344 & 0.220 & 0.320 & 0.203 & 0.312 \\
  \midrule 
  \multirow{4}{*}{\rotatebox{90}{Weather}} & 96 & \textcolor{red}{0.142} & \textcolor{red}{0.198} & \textcolor{blue}{\underline{0.147}} & \textcolor{blue}{\underline{0.198}} & 0.149 & 0.198 & 0.174 & 0.214 & 0.153 & 0.217 & 0.175 & 0.225 & 0.156 & 0.206 & 0.152 & 0.237 & 0.172 & 0.220 & 0.161 & 0.226 \\
  & 192 & \textcolor{red}{0.185} & \textcolor{red}{0.237} & \textcolor{blue}{\underline{0.194}} & \textcolor{blue}{\underline{0.238}} & 0.194 & 0.241 & 0.221 & 0.254 & 0.197 & 0.269 & 0.218 & 0.260 & 0.199 & 0.248 & 0.220 & 0.282 & 0.219 & 0.261 & 0.220 & 0.283 \\
  & 336 & \textcolor{red}{0.232} & \textcolor{red}{0.267} & \textcolor{blue}{\underline{0.235}} & \textcolor{blue}{\underline{0.277}} & 0.245 & 0.282 & 0.278 & 0.296 & 0.252 & 0.311 & 0.265 & 0.294 & 0.249 & 0.291 & 0.265 & 0.319 & 0.280 & 0.306 & 0.275 & 0.328 \\
  & 720 & \textcolor{red}{0.302} & \textcolor{red}{0.329} & \textcolor{blue}{\underline{0.308}} & \textcolor{blue}{\underline{0.331}} & 0.314 & 0.334 & 0.358 & 0.347 & 0.318 & 0.363 & 0.329 & 0.339 & 0.336 & 0.343 & 0.323 & 0.362 & 0.365 & 0.359 & 0.311 & 0.356 \\
  \midrule 
  \multirow{4}{*}{\rotatebox{90}{ILI}} & 24 & \textcolor{red}{1.320} & \textcolor{red}{0.703} & 1.324 & \textcolor{blue}{\underline{0.712}} & \textcolor{blue}{\underline{1.319}} & 0.754 & 2.207 & 1.032 & 3.040 & 1.186 & 4.337 & 1.507 & 1.472 & 0.798 & 2.215 & 1.081 & 2.317 & 0.934 & 2.684 & 1.112 \\
  & 36 & \textcolor{red}{1.180} & \textcolor{red}{0.764} & \textcolor{blue}{\underline{1.190}} & \textcolor{blue}{\underline{0.772}} & 1.430 & 0.834 & 1.934 & 0.951 & 3.356 & 1.230 & 4.205 & 1.481 & 1.435 & 0.745 & 1.963 & 0.963 & 1.972 & 0.920 & 2.507 & 1.013 \\
  & 48 & \textcolor{red}{1.395} & \textcolor{red}{0.706} & \textcolor{blue}{\underline{1.456}} & \textcolor{blue}{\underline{0.782}} & 1.553 & 0.815 & 2.127 & 1.004 & 3.441 & 1.223 & 4.257 & 1.484 & 1.474 & 0.822 & 2.130 & 1.024 & 2.238 & 0.940 & 2.423 & 1.012 \\
  & 60 & \textcolor{blue}{\underline{1.644}} & \textcolor{red}{0.782} & 1.652 & 0.796 & \textcolor{red}{1.470} & \textcolor{blue}{\underline{0.788}} & 2.298 & 0.998 & 3.608 & 1.302 & 4.278 & 1.487 & 1.839 & 0.912 & 2.368 & 1.096 & 2.027 & 0.928 & 2.653 & 1.085 \\
  \midrule 
  \multicolumn{2}{c}{1$^{st}$ Count} & \multicolumn{2}{c}{\textcolor{red}{40}} & \multicolumn{2}{c}{\textcolor{blue}{\underline{4}}} & \multicolumn{2}{c}{3} & \multicolumn{2}{c}{0} & \multicolumn{2}{c}{0} & \multicolumn{2}{c}{0} & \multicolumn{2}{c}{0} & \multicolumn{2}{c}{1} & \multicolumn{2}{c}{0} & \multicolumn{2}{c}{1} \\
  \bottomrule 
 \end{tabular}
 }
\end{table*}


The forecasting results are comprehensively presented in Table \ref{tab:res}. A clear trend is observed where the prediction accuracy tends to decline progressively as the forecast horizon length increases, reflecting the inherent challenges in long-term forecasting and the accumulation of uncertainty over time. Despite this general trend, MS-DFTVNet consistently demonstrates superior performance compared to other baseline models across diverse datasets such as Exchange, ETTm2, ETTh1, Electricity, Weather, and ILI. For example, on Exchange with a horizon of 336, MS-DFTVNet achieves an MSE of 0.271 versus 0.309 for TimesNet, while on ETTm2 at 720 it records 0.343 against 0.396 for iTransformer, corresponding to relative improvements of 12–15\%. Notably, the model achieves the lowest MSE and MAE in the majority of cases, highlighting its ability to capture both short-term fluctuations and long-term dependencies effectively. This advantage is particularly evident under extended forecasting horizons (e.g., 336 and 720 steps), where many Transformer-based models exhibit sharp performance degradation, while MS-DFTVNet maintains relatively stable accuracy. Such robustness suggests that the proposed multi-scale deformable convolution framework is well-suited for handling complex temporal patterns and cross-variable interactions.

\begin{figure}[ht]
\centering
\includegraphics[width=0.49\textwidth]{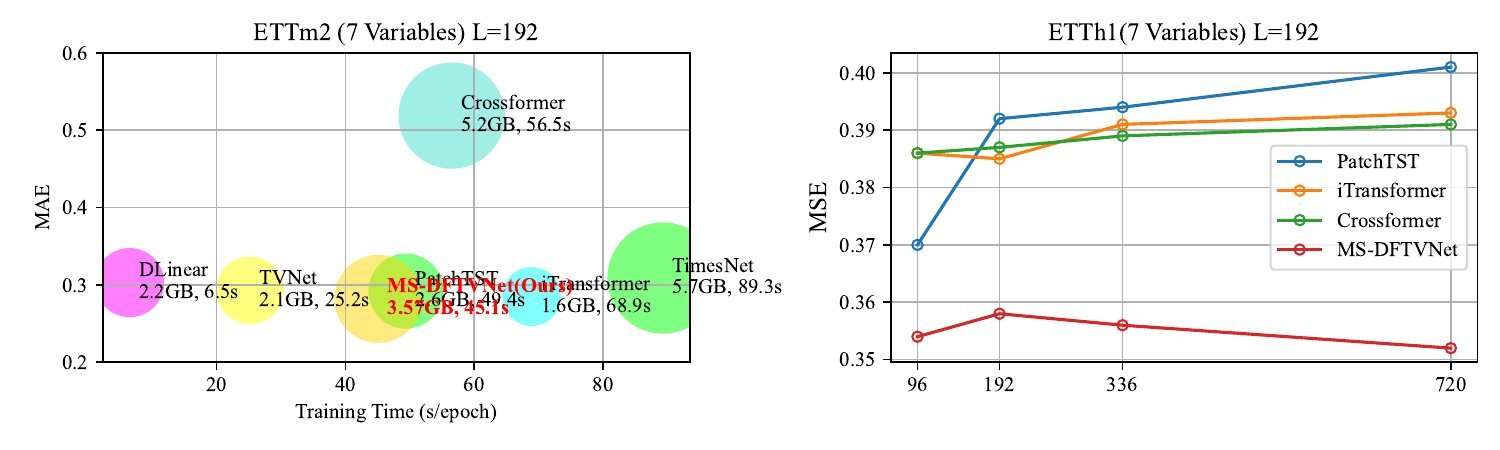}
\caption{Model efficiency comparison on ETTm2 with prediction length $L=192$, and performance comparison of models with different input lengths on ETTh1 ($L=192$).}
\label{fig:eff}
\end{figure}

Figure~\ref{fig:eff} presents the efficiency/memory and predictive behavior across models. The left panel shows that Transformer-based methods generally incur higher training time (s/epoch) and memory usage, whereas CNN models are overall lighter. Our MS-DFTVNet attains a moderate runtime (45.1 s/epoch) and low memory usage (3.57GB) while maintaining strong accuracy. The right panel further reveals that on ETTh1, as the forecasting horizon increases (96$\rightarrow$720), Transformer models’ MSE grows more sharply, whereas MS-DFTVNet consistently achieves the lowest error, highlighting its stability and practical value.
\begin{figure}
\centering
\includegraphics[width=0.9\linewidth]{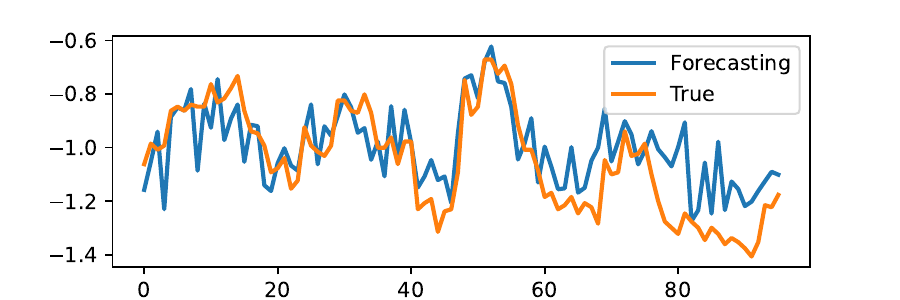}
\caption{Visualization of prediction results on ETTh1 with input length $L=96$.}
\label{fig:placeholder}
\end{figure}

\section{Conclusion}
In this paper, we propose MS-DFTVNet - a multi-scale 3D deformable convolutional neural network, which demonstrates outstanding performance in long-term time series prediction. This model can efficiently capture complex temporal patterns and cross-variable dependencies, achieving high accuracy in prediction tasks. Systematic evaluations on multiple public datasets further validate its state-of-the-art performance, establishing its potential as a powerful tool for time series prediction. These results highlight the application prospects of convolutional networks in time series.





\bibliographystyle{IEEEbib}
\bibliography{strings,refs}

\end{document}